\documentclass[conference]{IEEEtran}
\IEEEoverridecommandlockouts
\usepackage{cite}
\usepackage{tabularx}
\usepackage{graphicx}
\usepackage{subfig}
\usepackage{booktabs}      
\usepackage{colortbl}      
\usepackage{threeparttable}
\usepackage{url}
\usepackage{multirow}
\usepackage{amsmath,amssymb,amsfonts}
\usepackage{algorithmic}
\usepackage{textcomp}
\usepackage[export]{adjustbox}
\usepackage{xcolor}
\usepackage{hyperref}
\usepackage[utf8]{inputenc}
\usepackage{geometry}
\usepackage{chngcntr}
\usepackage{afterpage}
\usepackage{datetime}
\def\BibTeX{{\rm B\kern-.05em{\sc i\kern-.025em b}\kern-.08em
    T\kern-.1667em\lower.7ex\hbox{E}\kern-.125emX}}

\begin{document}

\title{Synthetic Art Generation and DeepFake Detection: A Study on Jamini Roy Inspired Dataset}

\author{
\IEEEauthorblockN{Kushal Agrawal}
\IEEEauthorblockA{\textit{m23csa011@iitj.ac.in}}
\and
\IEEEauthorblockN{Romi Banerjee}
\IEEEauthorblockA{\textit{romibanerjee@iitj.ac.in}}
}

\maketitle

\begin{abstract}
The intersection of generative AI and art is a fascinating area that brings both exciting opportunities and significant challenges, especially when it comes to identifying synthetic artworks. This study takes a unique approach by examining diffusion-based generative models in the context of Indian art, specifically focusing on the distinctive style of Jamini Roy.
To explore this, we fine-tuned Stable Diffusion 3 and used techniques like ControlNet and IPAdapter to generate realistic images. This allowed us to create a new dataset that includes both real and AI-generated artworks, which is essential for a detailed analysis of what these models can produce. We employed various qualitative and quantitative methods, such as Fourier domain assessments and autocorrelation metrics, to uncover subtle differences between synthetic images and authentic pieces.
A key takeaway from recent research is that existing methods for detecting deepfakes face considerable challenges, especially when the deepfakes are of high quality and tailored to specific cultural contexts. This highlights a critical gap in current detection technologies, particularly in light of the challenges identified above, where high-quality and culturally specific deepfakes are difficult to detect. 
This work not only sheds light on the increasing complexity of generative models but also sets a crucial foundation for future research aimed at effective detection of synthetic art.
\end{abstract}


\section{Introduction and Background}

With the rapid advancement of artificial intelligence, the realm of art generation has undergone profound transformation, utilizing various methods to create incredibly realistic and complex digital artwork. Among the most prominent methods are Generative Adversarial Networks (GANs), neural style transfer, and diffusion models. In particular, diffusion models are noted for their exceptional ability to infer a wide array of subjects and situations. These models can generate images that cover a vast range of topics, with limitations only set by the range of input words and initial images provided to the models~\cite{Lago_2022}.

In recent years, research on the detection of synthetic images has gained considerable momentum, especially after the emergence of highly realistic images generated by GANs. Some studies~\cite{Marra2018DetectionOG} have highlighted that the limitations of these generators in accurately replicating the high-level semantic features of natural images often result in visible artifacts, such as color anomalies and unnatural asymmetries.~\cite{Marra2018DetectionOG}


In response, some approaches~\cite{Marra2018DoGL,dzanic2020fourierspectrumdiscrepanciesdeep,corvi2022detectionsyntheticimagesgenerated,ricker2024detectiondiffusionmodeldeepfakes} now target low-level characteristics inherent in the underlying architecture of these models. These characteristics can be revealed by removing the high-level semantic details of the generated images~\cite{Marra2018DoGL}. The complex processing workflows involved in the generation of synthetic images often leave behind unique digital traces that differ significantly from those produced by traditional image capture devices. Previous research~\cite{dzanic2020fourierspectrumdiscrepanciesdeep,corvi2022detectionsyntheticimagesgenerated,ricker2024detectiondiffusionmodeldeepfakes} has explored these artifacts in the Fourier domain, focusing primarily on issues in GAN-generated images, such as distortions caused by upsampling during resolution scaling. However, relatively little attention has been paid to newer and promising architectures, such as those based on autoencoders and diffusion models.


This study systematically explores the detection of synthetic art in diffusion-based generative models, with a particular focus on Indian artworks inspired by Jamini Roy. This study specifically investigates two key questions: (i) whether diffusion models, like GANs, introduce identifiable artifacts in the Fourier domain and exhibit structured patterns in autocorrelation, and (ii) how effective state-of-the-art detectors are in recognizing these synthetic images.  
To provide context, the Literature Survey reviews existing deepfake detection methodologies, categorizing them into image-based and frequency-based approaches. This foundation informs the study’s approach and highlights the gaps in existing forensic detection methods.
To address these gaps, the Dataset Creation section details the construction of a novel dataset comprising both authentic Jamini Roy paintings and synthetic counterparts generated using a fine-tuned Stable Diffusion 3 model, augmented with ControlNet and IPAdapter. The Analysis employs qualitative, quantitative, and frequency-domain evaluations to uncover subtle generative artifacts that may serve as forensic markers. The Deepfake Detection Performance section benchmarks state-of-the-art models on this dataset, assessing their ability to differentiate real and AI-generated images. The Conclusion synthesizes key findings, while Challenges and Future Work propose enhancements to forensic methodologies to keep pace with advancements in generative AI. Finally, an appendix to Artifact Analysis provides deeper insights into the distinct spatial and frequency-based patterns that distinguish synthetic artworks.

\section{Literature survey}

In this section, a brief overview of deepfake detection methods is provided, organized into two main categories: image-based detection and frequency-based detection.

\subsection{Image-based Deepfake Detection}

Considerable work has been devoted to forgery detection, with numerous studies emphasizing the use of spatial information within images, such as irregular eye reflections. For example, the authors in~\cite{chai2020makesfakeimagesdetectable} employ restricted receptive fields to detect localized patches within images, highlighting the role of specific local features in making images detectable. Various methods~\cite{Chen2020SSDGANMT},~\cite{wang2023diffusiondblargescalepromptgallery} aim to enrich the diversity of training data by employing techniques such as data augmentation and adversarial training. In particular, recent work by the authors in~\cite{ojha2024universalfakeimagedetectors} employs feature maps as a general representation.

A more comprehensive analysis of synthetic image detectors~\cite{gragnaniello2021gangeneratedimageseasy} reveals that pre-training models on large datasets like ImageNet remains crucial because it significantly improves model robustness, uncertainty estimates, and performance on various tasks beyond traditional accuracy metrics. 

Given the rapid advancement of generative models, it is essential to develop detectors that can generalize across different generators, making this a key area of ongoing research~\cite{cozzolino2021universalganimagedetection}.Additionally, an emerging challenge lies in reverse engineering cultural semantics encoded in generated images. Unlike natural images, AI-generated artworks lack the deep-rooted contextual nuances present in traditional artworks, leading to stylistic inconsistencies that can serve as an implicit detection cue. Studies focusing on semantic inconsistencies, such as~\cite{tan2024frequencyawaredeepfakedetectionimproving}, suggest that certain cultural patterns embedded in artistic styles may be disrupted or misrepresented in AI-generated reproductions.

\subsection{Frequency-based Deepfake Detection}

In~\cite{Zhang2019DetectingAS}, evidence of frequency domain anomalies was first presented, manifesting as peaks in the Fourier domain. These peaks serve as distinguishing features for the development of forensic detectors. The traces are a result of the up-sampling operation in the decoder architecture of the generation process, which leads to a noticeable aliasing effect. Work in~\cite{frank2020leveragingfrequencyanalysisdeep} and~\cite{durall2020watchupconvolutioncnnbased}, which emphasize the significance of frequency-based artifacts in deepfake detection, have inspired subsequent image forgery detection methods to focus more on identifying distinctive patterns in the frequency domain. Durall's research~\cite{durall2020watchupconvolutioncnnbased} specifically demonstrated that GANs struggle to accurately replicate the spectral distribution of their training data. Notably, the generated images exhibit elevated magnitudes at high frequencies. This elevation in magnitude is observed to be approximately 0.4–0.5 in power spectrum units, marking a substantial increase compared to the lower values of around 0.1–0.2 found in the mid-frequency range (approximately 80–120 on the spatial frequency axis).

Authors in~\cite{li2021frequencyawarediscriminativefeaturelearning} developed a method that directs the detector to focus on high-frequency information, leveraging consequent feature representations across spatial and channel dimensions. This approach revealed that detectors lack proficiency in learning the frequency domain and tend to overfit to the artifacts present in the training data.

Recent studies~\cite{jeong2021bihpfbilateralhighpassfilters,corvi2023intriguingpropertiessyntheticimages,tan2024frequencyawaredeepfakedetectionimproving} have introduced frequency-based approaches for generalized detection. The BiHPF method~\cite{jeong2021bihpfbilateralhighpassfilters} enhances artifact magnitudes by applying dual high-pass filters. 

In contrast, whether images generated by diffusion models (DMs) exhibit grid-like frequency patterns appears to depend heavily on the specific model used~\cite{corvi2023intriguingpropertiessyntheticimages}, it allows for more nuanced forensic analysis of synthetic images, as different DM architectures may produce distinct artifacts, enabling better identification and classification of the source mode As a result, a key challenge is to develop universal detection methods that are effective across various generative models, particularly GANs and diffusion models. FreqNet~\cite{tan2024frequencyawaredeepfakedetectionimproving} addresses this challenge by focusing on high-frequency information, capturing source-agnostic features through both phase and amplitude spectra using FFT and iFFT. Similarly, Ojha et al.~\cite{ojha2024universalfakeimagedetectors} use a pre-trained vision transformer, CLIP-ViT, with an added classification layer instead of training directly on real and fake images. This approach avoids dependence on GAN-specific artifacts and improves generalization.

In the next section, we discuss the creation of a novel dataset and its significance in the context of deepfake detection for non-Western art styles.

\section{Dataset Creation and Novelty}  

The study of deepfake detection in the context of art has predominantly focused on Western art styles, with extensive datasets available for training and evaluation. However, to the best of our knowledge, such resources are scarce for Indian artists, such as Jamini Roy. To bridge this gap, a new dataset of Jamini Roy-inspired images was created using a fine-tuned Stable Diffusion 3 model, which combines both artificial and authentic representations of his style. This dataset not only addresses the lack of research resources to detect synthetic Indian artworks but also offers a unique foundation for studying the interactions between traditional artistic styles and modern generative techniques.  

Jamini Roy was chosen for this study due to his distinct departure from Western academic realism and his pioneering role in modern Indian art, making his style an ideal test case for deepfake detection in non-Western artistic traditions. Roy’s distinctive style~\cite{jamini_style} was built on a few foundational principles: he used flat, two-dimensional perspectives, inspired by patua paintings, rejecting Western notions of depth and realism. His palette was dominated by earthy colors—ochres, reds, blues, and greens—often derived from natural sources like tamarind seeds and powdered minerals. This conscious return to organic materials aligned with his commitment to folk traditions. Another hallmark of Roy’s style was his emphasis on bold, fluid line work, which created dynamic compositions and gave his figures a sense of rhythm and movement. His simplified facial features and almond-shaped eyes became iconic, capturing the essence of Indian folk and tribal art.  

By focusing on Jamini Roy, this study highlights the need for research in deepfake detection beyond widely studied Western art forms, ensuring a more inclusive and comprehensive approach to safeguarding artistic authenticity in the era of AI-generated art. The dataset is available at~\cite{jamini_dataset}.

The dataset creation process involved the following steps:

\subsection{Step 1: Image Extraction}
To build a robust dataset, real paintings by Jamini Roy were extracted from a comprehensive collection hosted on ArtNet~\cite{artnet}. A total of 770 images were obtained, selected for their diversity in style and subject matter, ensuring that the dataset captured the essence of his artistic expressions.

\subsection{Step 2: Selection of Hard Samples}
From these 770 images, 300 hard samples were chosen manually. Hard samples are defined as those exhibiting complex artistic elements, intricate details, or unique stylistic features characteristic of Jamini Roy's style. This intentional selection aimed to enhance the model's ability to learn nuanced representations, thereby facilitating the generation of artworks that closely mimic the original style.

\subsection{Step 3: Fine-tuning Stable Diffusion}
\begin{figure}[htbp]
        \centering
        \includegraphics[width=0.5\textwidth]{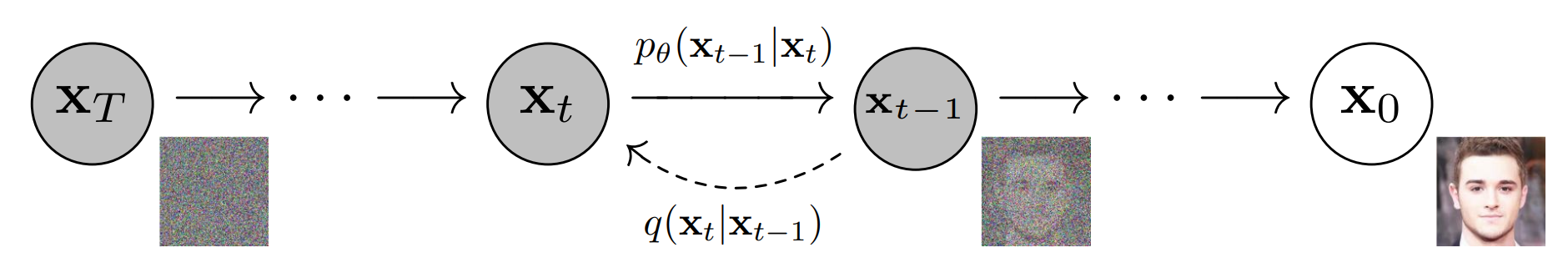}
        \caption{The directed graphical model}
        \label{fig:diffusion}
\end{figure}

In this study, Stable Diffusion 3~\cite{esser2024scalingrectifiedflowtransformers}, a state-of-the-art text-to-image diffusion model, was employed to generate synthetic images aligned with Jamini Roy’s artistic style. Stable Diffusion operates as a latent diffusion model, where denoising steps are performed in the latent space of a pre-trained autoencoder, significantly enhancing efficiency and fidelity. By iteratively refining noise over multiple steps, the model transforms a random latent vector into a high-quality image conditioned on input prompts. The architecture of Stable Diffusion is illustrated in Figure ~\ref{fig:diffusion}, adapted from~\cite{diffusion}.

This architecture enables effective fine-tuning for specialized styles with minimal computational cost, as the core denoising process remains in the compressed latent space. 
By working with a compressed latent representation, Stable Diffusion implicitly captures high-level features (low frequencies) and fine-grained details (high frequencies). This makes it particularly effective for generating art styles like Jamini Roy's, where capturing both overall composition and intricate stylistic details is critical. 

Following the selection of hard samples, the StabilityAI Stable Diffusion 3 medium diffusers~\cite{stability2024stablediffusion3} were fine-tuned using the DreamBooth tool to adapt the model for the specific artistic nuances of his style. DreamBooth facilitates domain-specific customization by enabling the model to internalize intricate patterns and details unique to the provided dataset. This step was crucial to ensure that the latent representations in the model's compressed space could accurately capture and replicate the subtle artistic features inherent in his works, enhancing the fidelity and stylistic consistency of the generated images. Fine-tuning is a critical process in transfer learning, where a pre-trained model is adapted to a new task or domain by further training on a specialized dataset. This process allows the model to leverage the general knowledge acquired during its initial training while refining its capabilities to generate outputs aligned with his artistic style. 
The model was trained for 28,000 steps on the 300 samples, with default parameters. Using default parameters for fine-tuning Stable Diffusion ensures that the training process remains consistent with the model's original configuration, which has been optimized through extensive testing. 
The prompt “art in the style of Jamini Roy” was carefully designed to guide the model during the image generation process, ensuring that the outputs captured the defining characteristics of Jamini Roy’s artistic style, such as bold lines, flat color palettes, and themes inspired by Indian folk culture.

\subsection{Step 4: Image Generation}
After fine-tuning, the model was utilized to generate images based on real paintings through image-to-image translation using the Automatic1111 tool. This approach allowed the model to synthesize new artworks that maintained the style of Jamini Roy while introducing novel elements like creative variations in composition, color schemes, and subject interpretations not found in the original works. Two types of experiments were conducted: 'Stable Diffusion 3' and 'Stable Diffusion 3 with ControlNet and IPAdapter'.

 \begin{table}[h!]
    \centering
    \caption{Stable Diffusion 3}
    \label{tab:no_controlnet}
    \resizebox{0.5\textwidth}{!}{%
    \begin{tabular}{|c|c|c|c|}
        \hline
        \textbf{Noise Level} & \textbf{Sample Image 1} & \textbf{Sample Image 2} & \textbf{Sample Image 3} \\
        \hline
        \centering Real Image & 
        \includegraphics[width=0.10\textwidth,valign=m]{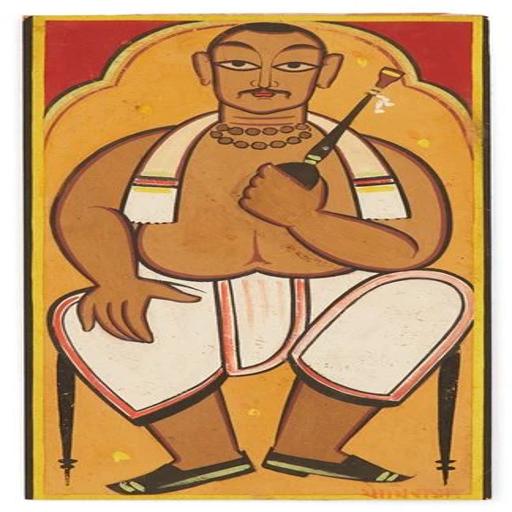} & 
        \includegraphics[width=0.10\textwidth,valign=m]{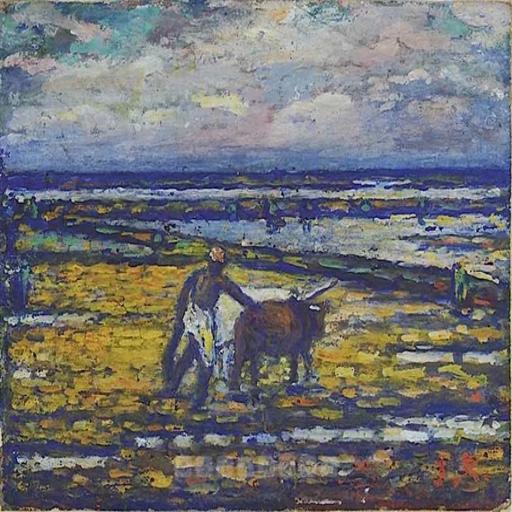} & 
        \includegraphics[width=0.10\textwidth,valign=m]{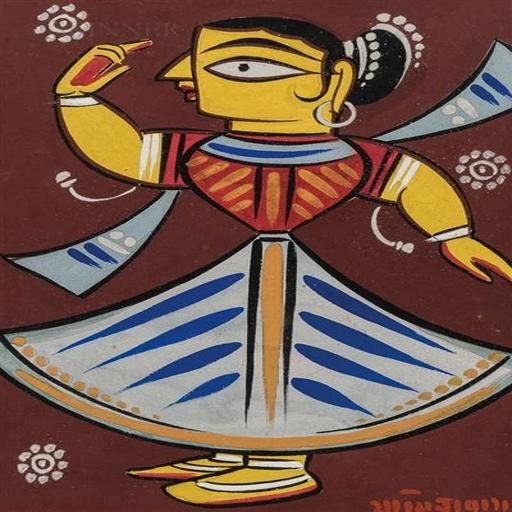}\\ 
        \hline
        \centering Noise 0.0 & 
        \includegraphics[width=0.10\textwidth,valign=m]{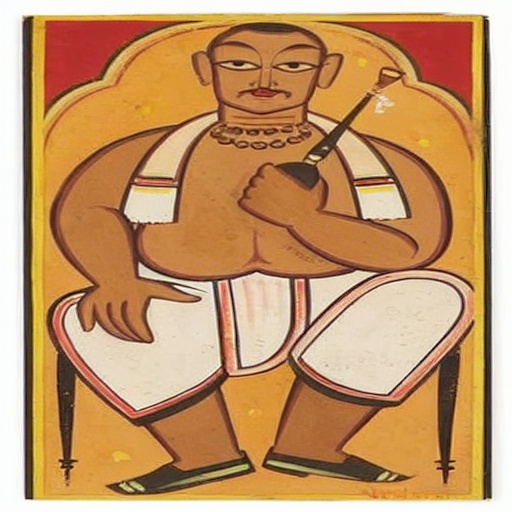} & 
        \includegraphics[width=0.10\textwidth,valign=m]{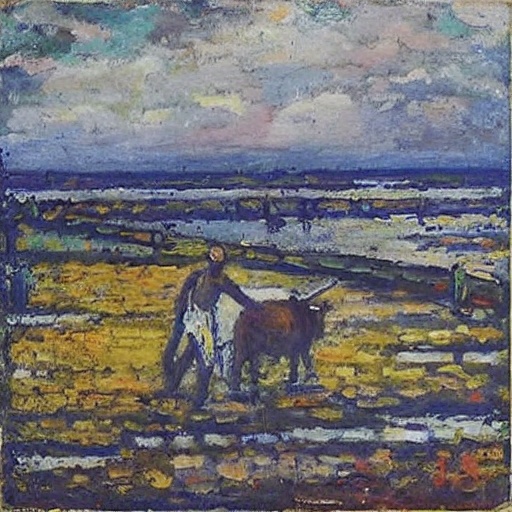} & 
        \includegraphics[width=0.10\textwidth,valign=m]{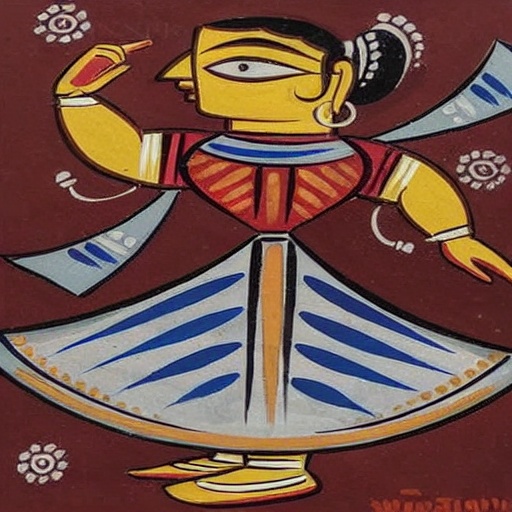}\\
        \hline
        \centering Noise 0.25 & 
        \includegraphics[width=0.10\textwidth,valign=m]{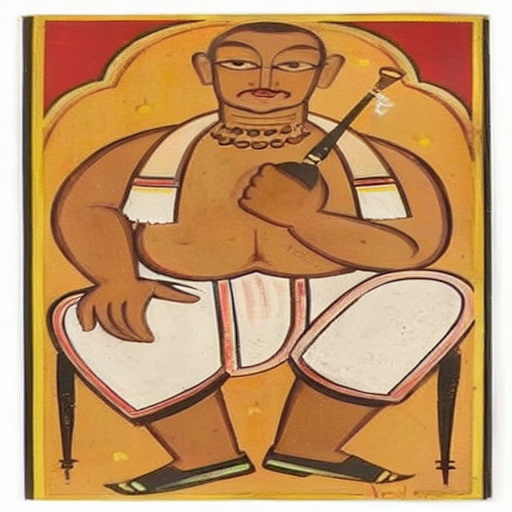} & 
        \includegraphics[width=0.10\textwidth,valign=m]{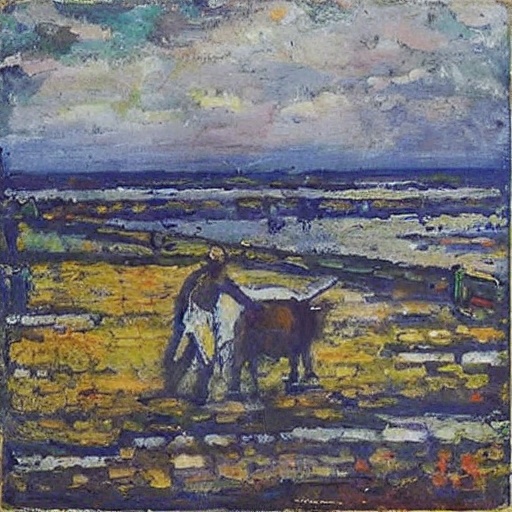} & 
        \includegraphics[width=0.10\textwidth,valign=m]{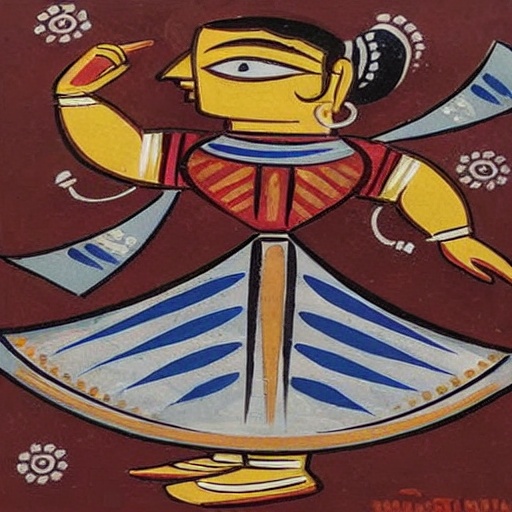}\\ 
        \hline
        \centering Noise 0.5 & 
        \includegraphics[width=0.10\textwidth,valign=m]{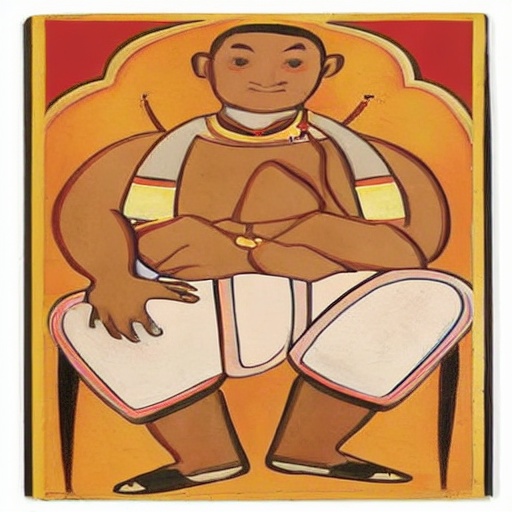} & 
        \includegraphics[width=0.10\textwidth,valign=m]{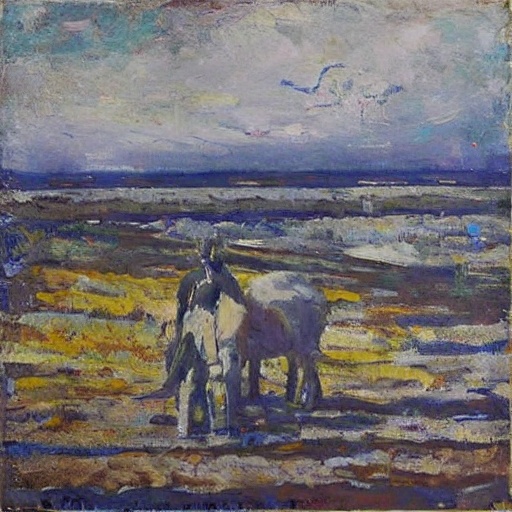} & 
        \includegraphics[width=0.10\textwidth,valign=m]{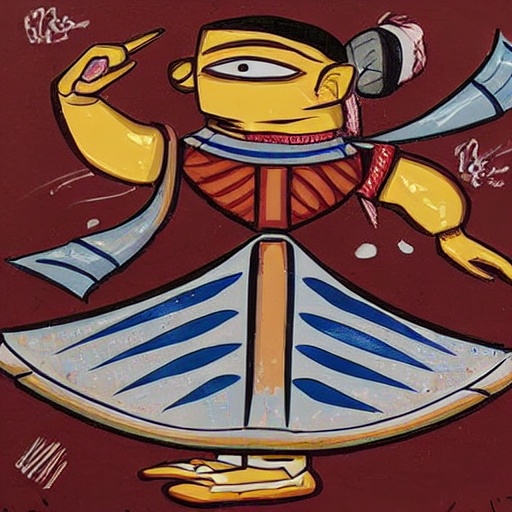} \\ 
        \hline
        \centering Noise 0.75 & 
        \includegraphics[width=0.10\textwidth,valign=m]{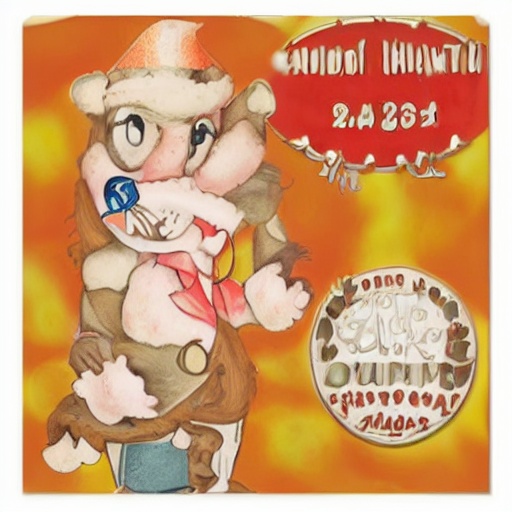} & 
        \includegraphics[width=0.10\textwidth,valign=m]{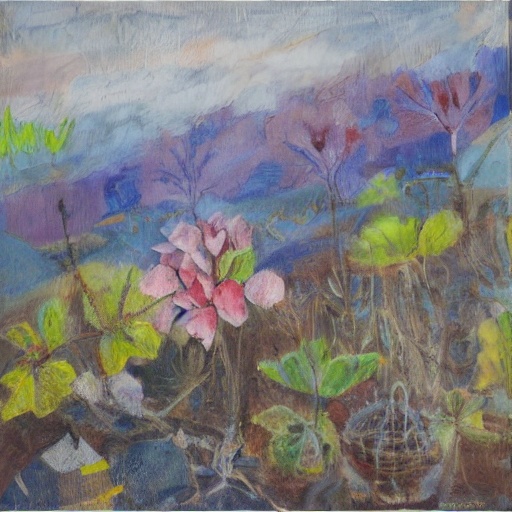} & 
        \includegraphics[width=0.10\textwidth,valign=m]{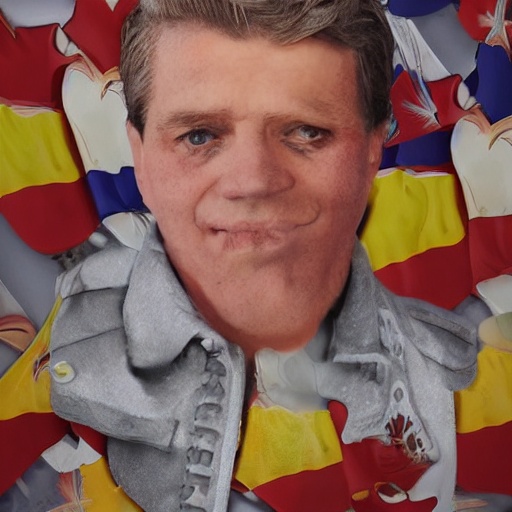}\\ 
        \hline
    \end{tabular}%
    }
\end{table}

\subsubsection{Stable Diffusion 3}

In this setup, the goal was to explore how the model performs under standard conditions without the additional constraints imposed by ControlNet. The configuration for Automatic1111 included:

\begin{itemize} \item \textbf{Basic Setup}: Sampling method: DPM++ 2M (default), Schedule type: Karras, Sampling steps: 100, Batch count/batch size: 1/1 (default), CFG scale: 7 (default) \end{itemize}

Synthetic artworks were generated with high similarity to Jamini Roy paintings, with 99.2\% similarity at Noise 0.0 based on mse. Similarity declined with increasing noise—96.8\% at Noise 0.25, 84.4\% at Noise 0.50, and 35.9\% at Noise 0.75 (0.0\% without ControlNet and IPAdapter) (see Table~\ref{tab:no_controlnet}) .The careful selection of noise levels allowed for a controlled examination of the model’s sensitivity to variations in input data. Noise 1.0 was not considered due to the emergence of extreme randomness and subtle signs of nudity in the generated images, which deviated from the intended artistic representation.
 
 \begin{table}[h!]
    \centering
    \caption{Stable Diffusion 3 with ControlNet}
    \label{tab:controlnet}
    \resizebox{0.5\textwidth}{!}{%
    \begin{tabular}{|c|c|c|c|}
        \hline
        \textbf{Noise Level} & \textbf{Sample Image 1} & \textbf{Sample Image 2} & \textbf{Sample Image 3} \\
        \hline
        \centering Real Image & 
        \includegraphics[width=0.10\textwidth,valign=m]{Images/Real/jamini_1.jpg} & 
        \includegraphics[width=0.10\textwidth,valign=m]{Images/Real/jamini_2.jpg} & 
        \includegraphics[width=0.10\textwidth,valign=m]{Images/Real/jamini_4.jpg}\\ 
        \hline
        \centering Noise 0.0 & 
        \includegraphics[width=0.10\textwidth,valign=m]{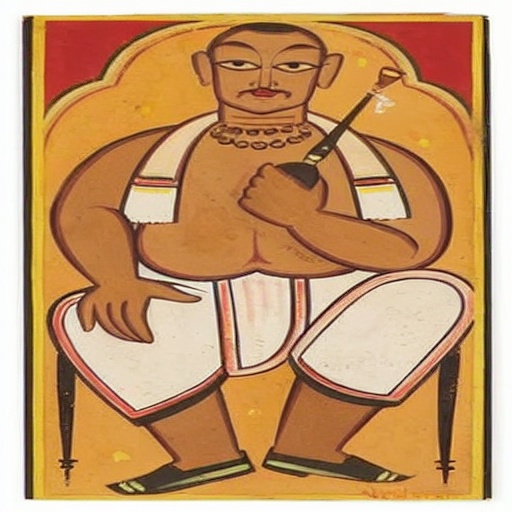} & 
        \includegraphics[width=0.10\textwidth,valign=m]{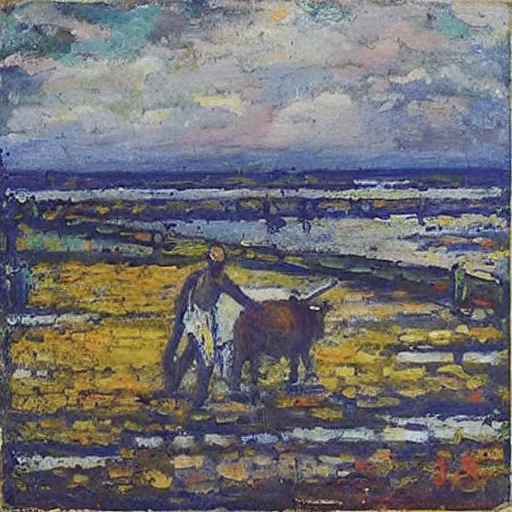} & 
        \includegraphics[width=0.10\textwidth,valign=m]{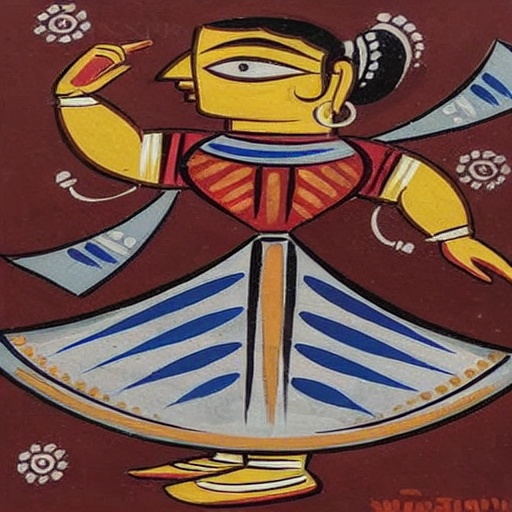}\\ 
        \hline
        \centering Noise 0.25 & 
        \includegraphics[width=0.10\textwidth,valign=m]{Images/controlnet/Noise_0.25/jamini_1.jpg} & 
        \includegraphics[width=0.10\textwidth,valign=m]{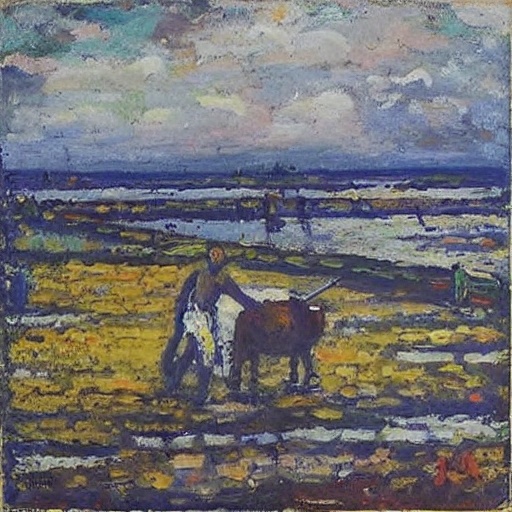} & 
        \includegraphics[width=0.10\textwidth,valign=m]{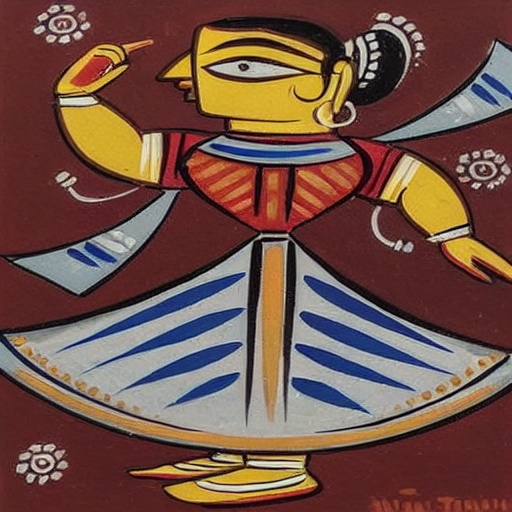}\\ 
        \hline
        \centering Noise 0.5 & 
        \includegraphics[width=0.10\textwidth,valign=m]{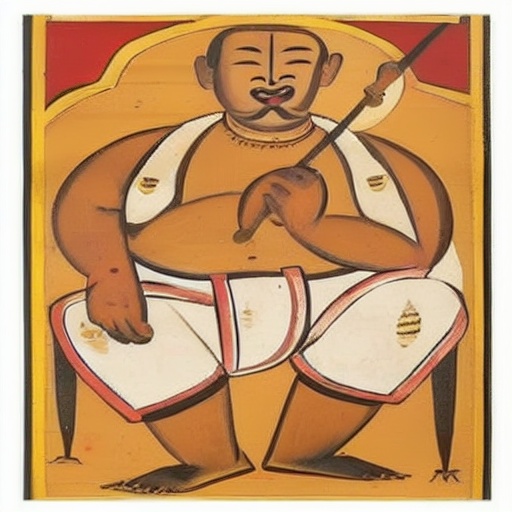} & 
        \includegraphics[width=0.10\textwidth,valign=m]{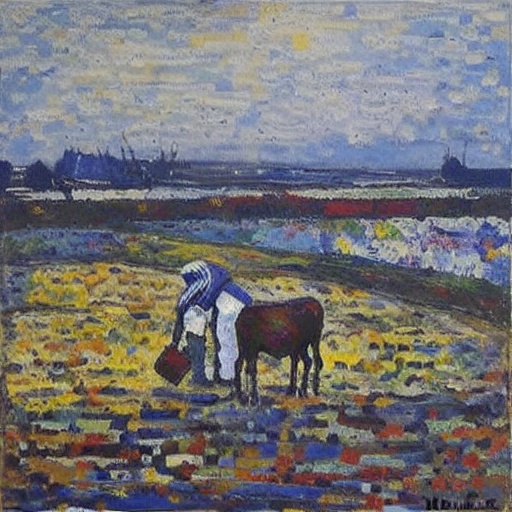} & 
        \includegraphics[width=0.10\textwidth,valign=m]{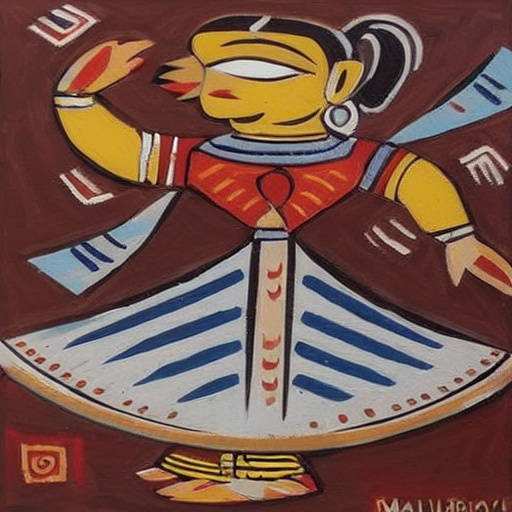} \\ 
        \hline
        \centering Noise 0.75 & 
        \includegraphics[width=0.10\textwidth,valign=m]{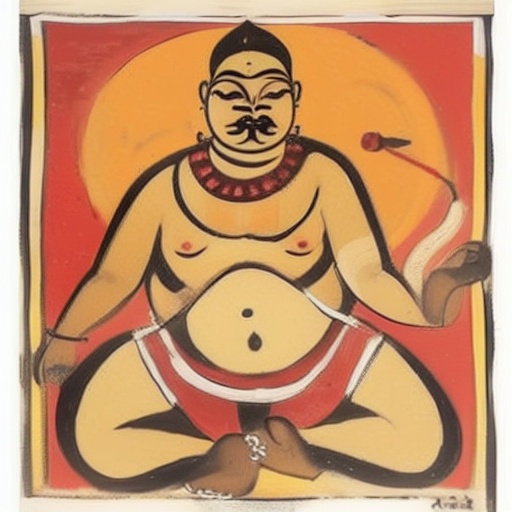} & 
        \includegraphics[width=0.10\textwidth,valign=m]{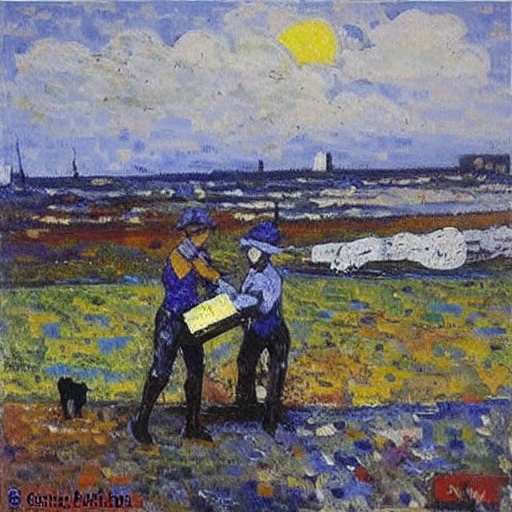} & 
        \includegraphics[width=0.10\textwidth,valign=m]{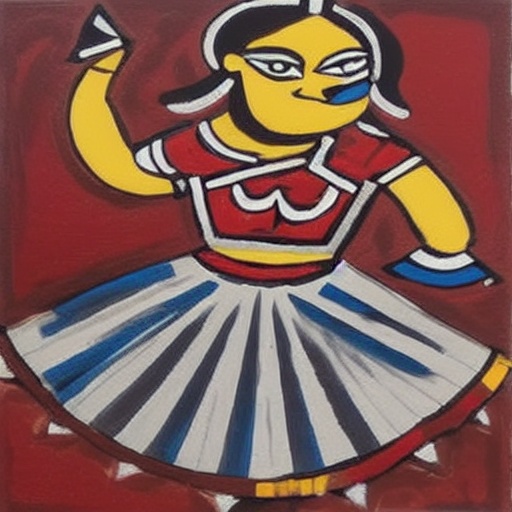}\\ 
        \hline
    \end{tabular} %
    }
\end{table}

\subsubsection{Stable Diffusion 3 with ControlNet and IPAdapter}

In this research, ControlNet and IP-Adapter, two advanced methods that enhance control over image generation in diffusion models, were leveraged. ControlNet enables fine-grained control by incorporating an external conditioning network that directs the diffusion process using auxiliary input, such as edge maps or pose data. This allows for targeted modifications without compromising the original model weights. The IP-Adapter further enhances this capability by integrating an intermediate perceptual module that aligns the generated images with specific visual attributes or styles, thereby improving both stylistic fidelity and accuracy.

The setup included:

\begin{itemize}
    \item \textbf{Basic Setup}: Sampling method: DPM++ 2M , Schedule type: Karras , Sampling steps: 100 , Batch count/batch size: 1/1 (default) , CFG scale: 7
    \item \textbf{ControlNet Unit Setup}:
    \begin{itemize}
        \item Control Type: IP Adapter , Preprocessor: ip-adapter-auto , Model: ip-adapter\_sd15 , Control weight: 1 (default) , Start Control Step: 0 (default) , Ending Control Step: 1 (default)
    \end{itemize}
\end{itemize}

Similar to the previous setup, the same noise categories were maintained, avoiding Noise 1.0 due to the appearance of subtle signs of nudity in the generated images, which did not align with Jamini Roy's artistic style. 

By combining these methods, nuanced control over style adherence and structural coherence was achieved (see Table~\ref{tab:controlnet}), demonstrating the effectiveness of ControlNet and IPAdapter in preserving key artistic features, even at higher noise levels. Quantitative metrics, as shown in Table~\ref{tab:quantitative_metrics}, highlight the ability of these methods to maintain stylistic fidelity while minimizing distortions in the generated images, especially when dealing with more challenging noise conditions.

\begin{table*}[htbp]
    \small
    \setlength{\tabcolsep}{4pt}  
    \begin{tabular}{|p{0.1\textwidth}|p{0.45\textwidth}|p{0.45\textwidth}|}
        \hline
        \multicolumn{3}{|c|}{\textbf{Comparison of Noise Levels in Stable Diffusion Models}} \\
        \hline
        \textbf{Noise Lvl} & \textbf{Stable Diffusion 3} & \textbf{Stable Diffusion 3 with ControlNet and IPAdapter} \\
        \hline
        0.0 & 
        Images closely resembled the original artworks, with a high degree of accuracy in features such as clothing, and facial expressions. The absence of apparent artifacts, and uniform texture and color proved to be significant obstacles for detection systems. & 
        Images exhibited characteristics similar to the original pieces of art, including intricate detailing and lighting. This made the chances of detecting deepfakes almost impossible. \\
        \hline
        0.25 & 
        Certain features of the face or clothing style began to show clues. However, the textures became smoother, and the composition remained relatively intact with minimal blurring. & 
        The generated images were nearly identical to the originals, with only minor variations, even in the presence of more noise. \\
        \hline
        0.5 & 
        Major stylistic differences began to emerge, distorting the uniqueness of the images. Although the foundational proportions were maintained. & 
        The same poses were generally maintained, the markers for identification included less detailed textures and moderate lighting. \\
        \hline
        0.75 & 
        Significant deviations became evident at this stage, with notable alterations in poses, facial features, and attire.  & 
        The same poses were generally maintained, but variations in facial expressions, hairstyles, and clothing added a modern twist. \\
        \hline
    \end{tabular}
    \caption{Comparative analysis of noise levels between Stable Diffusion 3 and Stable Diffusion 3 with ControlNet and IPAdapter}
    \label{tab:noise-comparison}
\end{table*}

\section{Analysis}

\subsection{Qualitative Analysis}

In the qualitative analysis, a detailed evaluation of images generated by the Stable Diffusion model was presented, focusing on how faithfully they captured the essence of Jamini Roy's original paintings. The challenges of detecting these images as deepfakes were explored, particularly as noise levels were manipulated (see Table~\ref{tab:noise-comparison}).

\subsection{Quantitative Analysis}

To supplement the qualitative observations, quantitative analysis was conducted, evaluating the fidelity of AI-generated images against authentic Jamini Roy paintings. The evaluation utilized various metrics, including Mean Squared Error (MSE), Structural Similarity Index (SSIM), Peak Signal-to-Noise Ratio (PSNR), and Histogram Correlation. Each metric provided insights into different aspects of similarity and structural integrity between the generated images and authentic artwork.

\begin{figure*}[h!]
    \centering
    \includegraphics[width=0.9\textwidth]{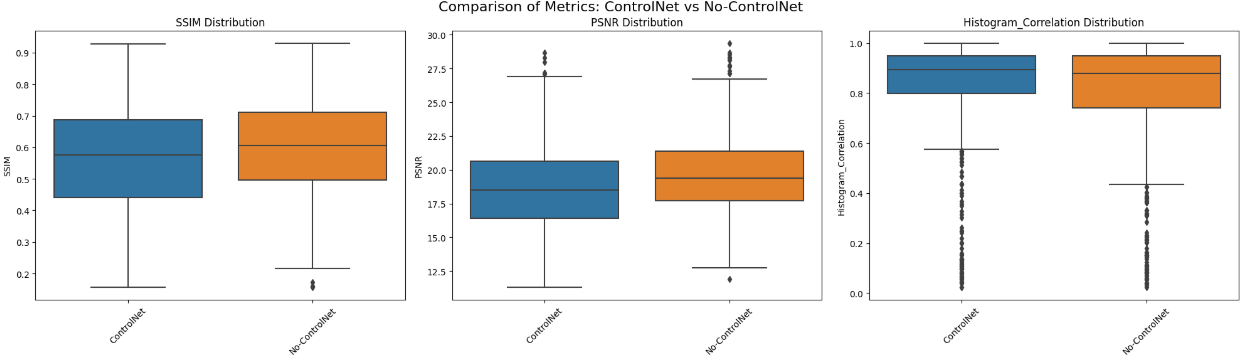}
    \caption{Comparison of Metrics: The SSIM, PSNR, and Histogram Correlation distributions are shown for images generated with and without ControlNet and IPAdapter}
    \label{fig:boxplots}
\end{figure*}

\begin{itemize}
    \item \textbf{Mean Squared Error (MSE):} Fig.~\ref{fig:dissimilarity} illustrates the MSE values across varying noise levels for images generated with and without ControlNet and IPAdapter. The results indicate that dissimilarity increases with noise, with non-ControlNet images exhibiting sharper increases in MSE at higher noise levels. ControlNet with IPAdpater retains stylistic consistency, resulting in lower MSE values at noise levels 0.5 and 0.75.
    
    \begin{figure}[htbp]
        \centering
        \includegraphics[width=0.5\textwidth]{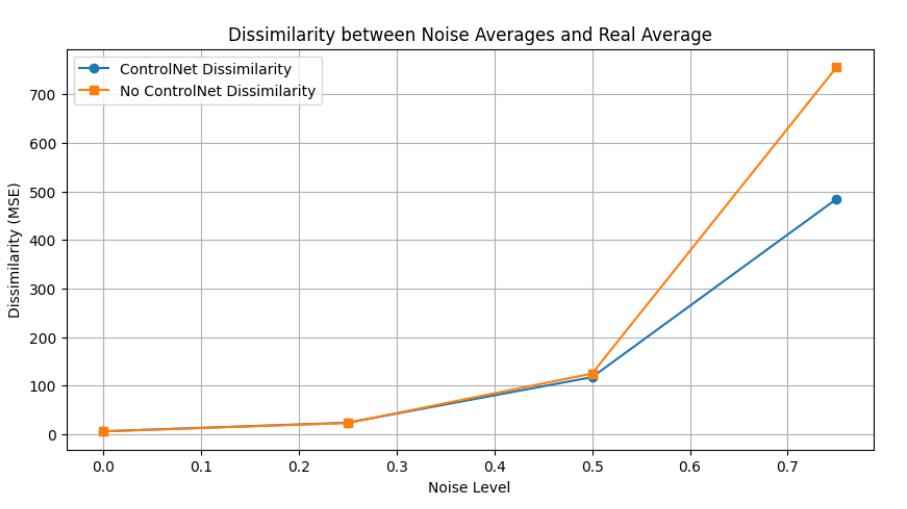}
        \caption{Mean Squared Error (MSE) analysis demonstrating the relationship between noise levels and dissimilarity in images generated with and without ControlNet.}
        \label{fig:dissimilarity}
    \end{figure}
    
    \item \textbf{Structural Similarity Index Measure (SSIM):} (see Fig. \ref{fig:boxplots}) While both methods show similar SSIM distributions, ControlNet achieves a comparable mean of 0.7287 (versus 0.7318 for non-ControlNet) with nearly identical standard deviation (0.1216 vs 0.1215). This demonstrates ControlNet's ability to maintain high structural fidelity to real images while providing additional control over the generation process.
    \item \textbf{Peak Signal-to-Noise Ratio (PSNR):} (see Fig. \ref{fig:boxplots}) ControlNet produces images with a strong PSNR of 25.16 dB, close to non-ControlNet's 25.35 dB. The minimal difference of 0.19 dB is negligible in practical applications, showing that ControlNet achieves this comparable quality while offering superior control over the output.
    \item \textbf{Histogram Correlation:} (see Fig. \ref{fig:boxplots}) Both methods demonstrate excellent color distribution alignment with real paintings, with ControlNet achieving a robust correlation of 0.8615 compared to 0.8761 for non-ControlNet. 
\end{itemize}

Table~\ref{tab:quantitative_metrics} summarizes these quantitative results, enabling a side-by-side comparison between with ControlNet and without ControlNet methods. These findings, alongside Fig. ~\ref{fig:dissimilarity}, reinforces ControlNet's effectiveness in preserving artistic features, even at higher noise levels.

\begin{table}[h!]
    \centering
    \caption{Quantitative Metrics Comparison between ControlNet and No-ControlNet Methods}
    \label{tab:quantitative_metrics}
    \begin{tabular}{lcc}
        \hline
        \textbf{Metric} & \textbf{ControlNet} & \textbf{No-ControlNet} \\
        \hline
        SSIM & 0.7287 $\pm$ 0.1216 & 0.7318 $\pm$ 0.1215 \\
        PSNR & 25.16 dB $\pm$ 3.12 dB & 25.35 dB $\pm$ 3.19 dB \\
        Correlation & 0.8615 $\pm$ 0.2252 & 0.8761 $\pm$ 0.2201 \\
        \hline
    \end{tabular}
\end{table}

\subsection{Frequency Domain Analysis}

A frequency domain analysis was conducted to differentiate between real Jamini Roy paintings and AI-generated images by examining their average power spectra (see Table~\ref{tab:frequency_domain}). Authentic paintings exhibit a distinctive cross-shaped pattern in the high-frequency domain, reflecting the artist's intricate brushwork and balanced complexity across frequency scales, with pronounced energy along both horizontal and vertical axes. In contrast, images generated with ControlNet display a 95.3\% similar power spectrum based on mse, but with diminished intensity and fewer high-frequency details, indicating a loss of fine textures. Images generated without ControlNet present the most uniform power spectrum, characterized by minimal high-frequency components and substantial over-smoothing.

\begin{table}[h!]
    \centering
    \caption{Frequency Domain Analysis}
    \label{tab:frequency_domain}
    \begin{tabular}{|c|c|c|}
        \hline
        \textbf{Real} & \textbf{With IPAdapter} & \textbf{Without IPAdapter} \\
        \hline
        \centering 
        \includegraphics[width=0.125\textwidth,valign=m]{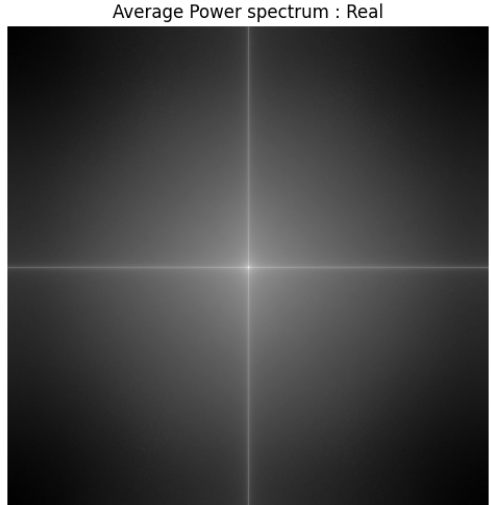} & 
        \includegraphics[width=0.125\textwidth,valign=m]{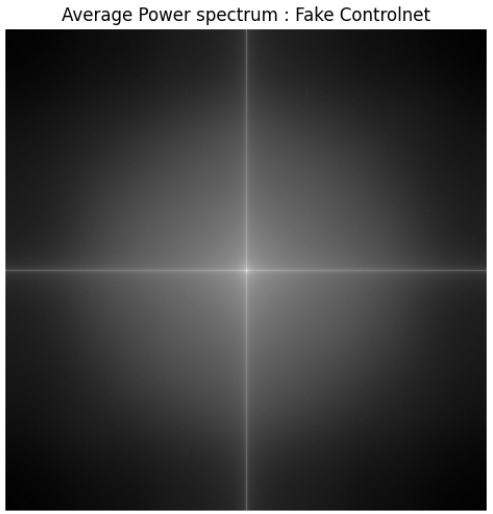} & 
        \includegraphics[width=0.125\textwidth,valign=m]{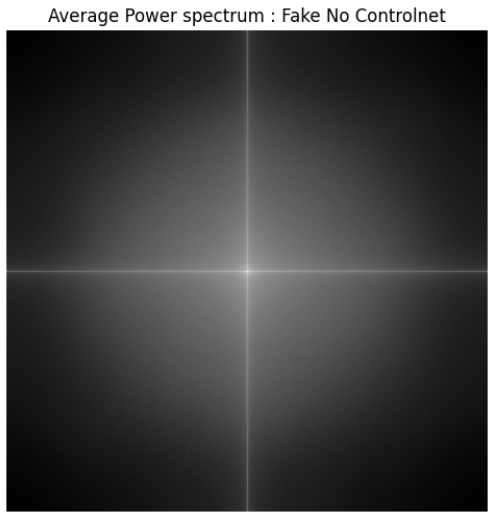} \\
        \hline
    \end{tabular}
\end{table}

\subsection{Frequency Domain Analysis (Single Image)}

Frequency domain analysis was performed on three versions of a Jamini Roy painting: the original artwork, one version generated by Stable Diffusion 3, and another version generated by Stable Diffusion 3 with ControlNet and IPAdapter (both AI versions using noise level 0.0). The analysis examined both power and phase spectra to identify distinct spectral characteristics across each type.

\subsubsection{Power Spectrum Analysis}
\begin{itemize}
    \item \textbf{Original Artwork}: Displays a sharp cross-pattern with strong radial components, high mid-frequency intensity, and asymmetric energy distribution, particularly along diagonal axes, signifying intricate artistic detail (see Fig.~\ref{fig:original_power_spectrum}).
    \item \textbf{Without ControlNet Generated Image}: Exhibits the most diffused power spectrum, with diminished cross-pattern contrast and an isotropic energy distribution, lacking the fine-grained frequency elements found in the original (see Fig.~\ref{fig:nocontrolnet_power_spectrum}). Quantitative metrics show a cross pattern strength of 96.4\%, mid-frequency intensity of 95.3\%, and asymmetry of 88.8\%.
    \item \textbf{With ControlNet Generated Image}: Preserves 95.3\% similar power spectrum structure based on mse with reduced clarity and slightly blurred high-frequency content, showing a more uniform energy distribution across frequencies while retaining major structural frequencies (see Fig.~\ref{fig:controlnet_power_spectrum}). Quantitative analysis reveals a cross pattern strength of 96.6\%, mid-frequency intensity of 95.1\%, and asymmetry of 85.7\%.
\end{itemize}

\subsubsection{Phase Spectrum Analysis}
\begin{figure*}[h!]
    \centering
    \includegraphics[width=0.5\textwidth]{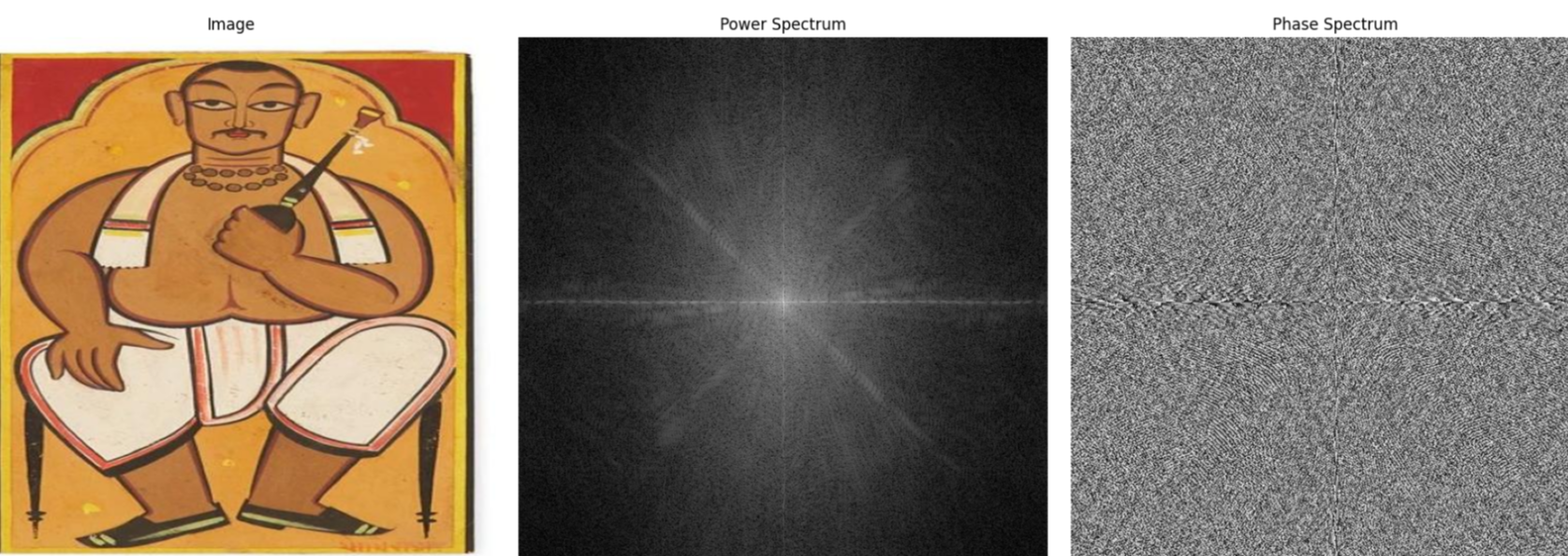}
    \caption{Power Spectrum And Phase Spectrum of the Original Artwork}
    \label{fig:original_power_spectrum}
\end{figure*}

\begin{figure*}[h!]
    \centering
    \includegraphics[width=0.5\textwidth]{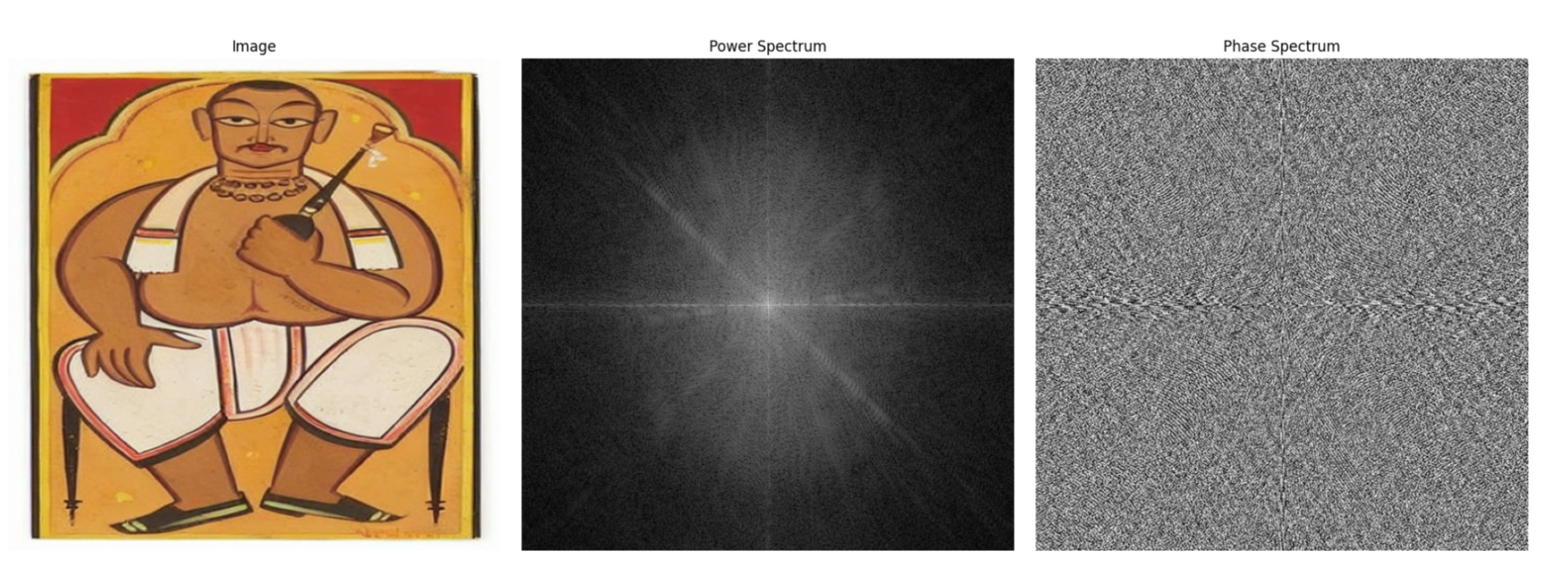}
    \caption{Power Spectrum And Phase Spectrum of Stable Diffusion Generated Image without ControlNet and IPAdapter}
    \label{fig:nocontrolnet_power_spectrum}
\end{figure*}

\begin{figure*}[h!]
    \centering
    \includegraphics[width=0.5\textwidth]{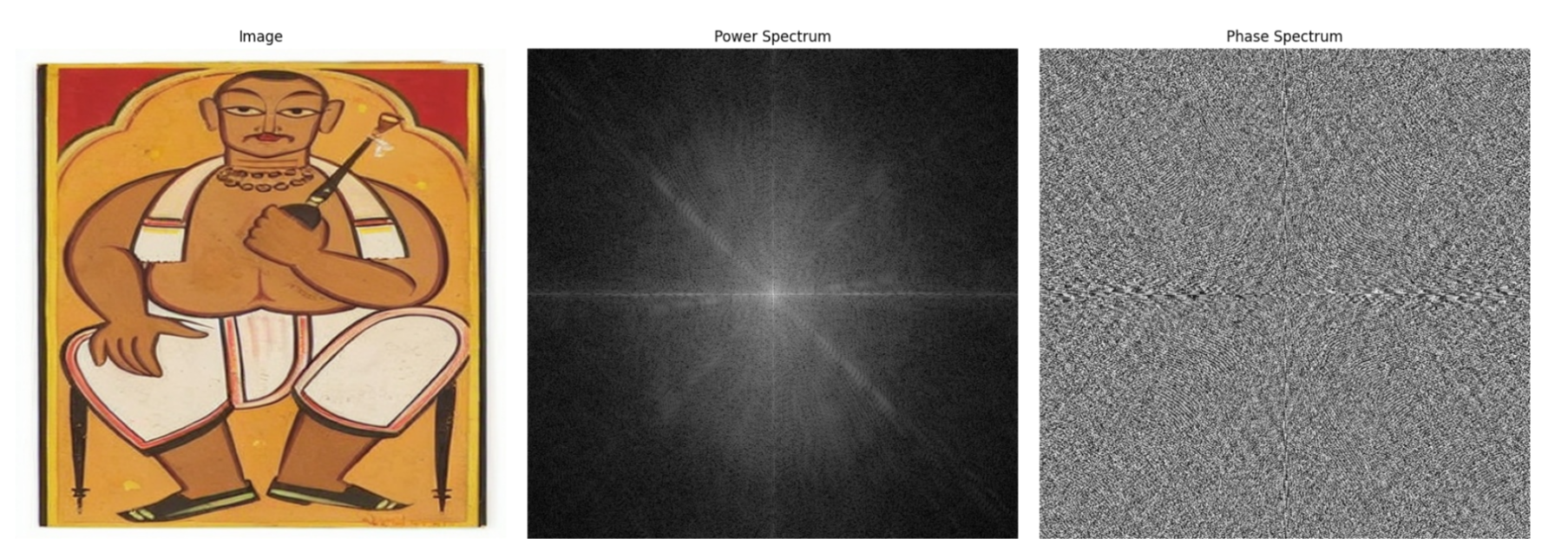}
    \caption{Power Spectrum And Phase Spectrum of Stable Diffusion Generated Image with ControlNet and IPAdapter}
    \label{fig:controlnet_power_spectrum}
\end{figure*}

\begin{itemize}
    \item \textbf{Original Artwork}: Exhibits complex, non-uniform phase patterns with clear structural organization, high local coherence, and sharp phase transitions (see Fig.~\ref{fig:original_power_spectrum}).
    \item \textbf{Without ControlNet Generated Image}: Shows the most uniform phase distribution, with regularized patterns and minimal phase transitions, indicating lower phase complexity compared to the original (see Fig.~\ref{fig:nocontrolnet_power_spectrum}). Quantitative metrics reveal 99.9\% local coherence, 98.1\% phase transitions, 100\% pattern complexity, and 99.7\% structural organization.
    \item \textbf{With ControlNet Generated Image}: Displays a phase structure that is 15.2\% similar to the original based on mse, with moderate coherence, but a reduced complexity in finer details (see Fig.~\ref{fig:controlnet_power_spectrum}). Quantitative analysis indicates 100\% local coherence, 97.9\% phase transitions, 100\% pattern complexity, and 98\% structural organization.
\end{itemize}

The following section focuses on evaluating the performance of existing state-of-the-art deepfake detection models using the novel Jamini Roy-inspired dataset, highlighting their effectiveness in identifying synthetic Indian artwork under varying conditions.

\section{DeepFake Detection Performance}

The detection performance of three state-of-the-art models was evaluated on the novel Jamini Roy-inspired dataset, providing a critical benchmark for deepfake detection in the context of Indian artwork. Given the unique challenges posed by synthetic art and the limited resources available for studying deepfake detection in non-Western art, this study fills a significant gap by examining the robustness of these models against various noise levels in a dataset that closely mimics his distinctive style.

The dataset used in the evaluation contained 770 images per category, with both real Jamini Roy-inspired artworks and synthetic images generated via a fine-tuned Stable Diffusion model. The synthetic images were categorized by four noise levels (0, 0.25, 0.5, and 0.75) and further divided into another set created with the same model, enhanced using ControlNet and IPAdapter. 
This structure guarantees a comprehensive assessment, especially in evaluating model resilience against subtle visual artifacts that can be influenced by different noise levels and control mechanisms.


For evaluating detection performance on the generated dataset, three state-of-the-art approaches were selected to address the unique challenges posed by synthetic Indian artwork. Model 1, the frequency-aware deepfake detection model by Tan et al. (2024), employs frequency-space learning to detect common artifacts in synthetic images, making it well-suited for analyzing diverse datasets with distinct artistic styles. Model 2, developed by Corvi et al. (2023)~\cite{corvi2022detectionsyntheticimagesgenerated}, focuses specifically on artifacts generated by diffusion models, aligning closely with the synthetic images created for this study. Lastly, Model 3, the universal fake image detector by Ojha et al. (2024), leverages feature maps as a general representation, offering versatility across generative models, although its performance may vary depending on the noise levels and control mechanisms influencing the synthetic image artifacts.

By employing these models, this research provides a critical benchmark for evaluating the robustness of deepfake detection techniques against varying noise levels and subtle visual artifacts, especially in the context of preserving the integrity of Indian artwork.

\subsection{Experimental Setup and Results}

\begin{table*}[htbp]
\centering
\caption{Validation Accuracy (\%) for DeepFake Detection Models across Dataset Categories}
\label{tab:results}
\resizebox{\textwidth}{!}{%
\begin{tabular}{lccc}
\hline
Dataset Generation Type & Model 1~\cite{tan2024frequencyawaredeepfakedetectionimproving} & Model 2~\cite{corvi2022detectionsyntheticimagesgenerated} & Model 3~\cite{ojha2024universalfakeimagedetectors} \\
\hline
Without ControlNet (Noise 0.0) & 54.6 & \underline{76.1} & 53.5 \\
Without ControlNet (Noise 0.25) & 57.1 & \underline{72.2} & 53.9 \\
Without ControlNet (Noise 0.50) & 68.4 & \underline{80.2} & 69.4\\
Without ControlNet (Noise 0.75) & 74.2 & 93.0 & \underline{94.9} \\
\hline
With ControlNet (Noise 0.0) & 58.6 & \underline{75.9} & 57.6 \\
With ControlNet (Noise 0.25) & 58.1 & \underline{74.7} & 53.0 \\
With ControlNet (Noise 0.50) & 58.6 & \underline{83.4} & 75.9 \\
With ControlNet (Noise 0.75) & 86.4 & \underline{91.8} & 88.5 \\
\hline
\end{tabular}%
}
\end{table*}

Each model was trained on the dataset for 50 epochs with default settings, resulting in validation accuracy over eight separate categories (with and without ControlNet and IPAdapter and four noise levels each).
Detailed results are presented in Table~\ref{tab:results}.
Model 2, which was designed specifically for applications in diffusion models, clearly outperformed the other models in majority of the scenarios, particularly in the high-noise case. Its performance was strong in detecting artifacts in synthetically generated images by diffusion models, which regularized its noise robustness across the varying noise levels in the dataset. Model 3 showed the best performance in fake image classification for maximum noise, but its general nature and flexibility resulted in fluctuating performances in different scenarios, particularly in the case of ControlNet, which improved image quality and reduced the extent of artifacts. Even so, it performed well, especially at the higher end of the noise levels. Conversely, Model 1, which employed a frequency-aware approach to detection, struggled with images enhanced by ControlNet and showed no significant increase in detection accuracy, even after high levels of noise were introduced. While this model performed well in noisy situations, several alterations and modifications may be needed for it to function effectively in control-dominated environments that introduce lower-level artifacts.

These varying performances highlight not only the challenges faced by current deepfake detection models when applied to synthetic Indian artwork but also emphasize the need for more adaptive, specialized approaches.

\section{Conclusion}


This research addresses the critical challenge of detecting synthetic images generated by diffusion models, with a particular focus on the underrepresented domain of Indian art. Using Jamini Roy's style as a case study, a curated dataset of real and AI-generated images explores deepfake detection in this cultural context. The findings show that while ControlNet and IPAdapter enhance image quality, they also obscure common artifacts, making detection harder. Despite these advances, the analysis showed in figure~\ref{fig:artifacts_analysis} demonstrates that certain noise residual artifacts remain detectable, particularly in the frequency domain.

The experimental evaluation of state-of-the-art deepfake detectors across various noise levels highlights the complexity of the problem. Detector performance varied significantly, reflecting the diverse forensic cues utilized by different models. These findings suggest that current deepfake detection methods are still inadequate to handle the unique characteristics of diffusion-based fakes, especially in the context of art forms with distinct stylistic elements.

\section{Challenges and Future Work}

The limitations of available datasets significantly impacted the performance of state-of-the-art detection models, pointing to a pressing need for more robust data collection efforts. One promising direction for future research is the application of few-shot learning techniques to address the scarcity of training data and improve model generalization with limited examples. Additionally, future work should focus on refining detection methodologies to better capture the subtle and model-specific artifacts inherent in diffusion-based synthetic images.

As art styles evolve over time, continual learning-based models will be essential to keep pace with these changes and maintain the relevance of detection tools. Furthermore, while current detection models are heavily reliant on frequency domain artifacts, future advancements in generative models may minimize or eliminate such artifacts, rendering existing methods obsolete. To counter this, the development of hybrid detection approaches that integrate both spatial and frequency domain features is advocated, ensuring more robust detection across a broader range of synthetic images.

Lastly, testing this study on works of other Indian artists, beyond the current dataset, could provide valuable insights into the generalizability of the detection methodologies. This exploration could uncover unique challenges presented by diverse styles and broaden the applicability of these models.


\bibliographystyle{IEEEtran}
\addcontentsline{toc}{section}{References}
\bibliography{refs}

\begin{thebibliography}{10}
\providecommand{\url}[1]{#1}
\csname url@samestyle\endcsname
\providecommand{\newblock}{\relax}
\providecommand{\bibinfo}[2]{#2}
\providecommand{\BIBentrySTDinterwordspacing}{\spaceskip=0pt\relax}
\providecommand{\BIBentryALTinterwordstretchfactor}{4}
\providecommand{\BIBentryALTinterwordspacing}{\spaceskip=\fontdimen2\font plus
\BIBentryALTinterwordstretchfactor\fontdimen3\font minus \fontdimen4\font\relax}
\providecommand{\BIBforeignlanguage}[2]{{%
\expandafter\ifx\csname l@#1\endcsname\relax
\typeout{** WARNING: IEEEtran.bst: No hyphenation pattern has been}%
\typeout{** loaded for the language `#1'. Using the pattern for}%
\typeout{** the default language instead.}%
\else
\language=\csname l@#1\endcsname
\fi
#2}}
\providecommand{\BIBdecl}{\relax}
\BIBdecl

\bibitem{Lago_2022}
\BIBentryALTinterwordspacing
F.~Lago, C.~Pasquini, R.~Bohme, H.~Dumont, V.~Goffaux, and G.~Boato, ``More real than real: A study on human visual perception of synthetic faces [applications corner],'' \emph{IEEE Signal Processing Magazine}, vol.~39, no.~1, p. 109–116, Jan. 2022. [Online]. Available: \url{http://dx.doi.org/10.1109/MSP.2021.3120982}
\BIBentrySTDinterwordspacing

\bibitem{Marra2018DetectionOG}
\BIBentryALTinterwordspacing
F.~Marra, D.~Gragnaniello, D.~Cozzolino, and L.~Verdoliva, ``Detection of gan-generated fake images over social networks,'' \emph{2018 IEEE Conference on Multimedia Information Processing and Retrieval (MIPR)}, pp. 384--389, 2018. [Online]. Available: \url{https://api.semanticscholar.org/CorpusID:49539556}
\BIBentrySTDinterwordspacing

\bibitem{Marra2018DoGL}
\BIBentryALTinterwordspacing
F.~Marra, D.~Gragnaniello, L.~Verdoliva, and G.~Poggi, ``Do gans leave artificial fingerprints?'' \emph{2019 IEEE Conference on Multimedia Information Processing and Retrieval (MIPR)}, pp. 506--511, 2018. [Online]. Available: \url{https://api.semanticscholar.org/CorpusID:57189570}
\BIBentrySTDinterwordspacing

\bibitem{dzanic2020fourierspectrumdiscrepanciesdeep}
\BIBentryALTinterwordspacing
T.~Dzanic, K.~Shah, and F.~Witherden, ``Fourier spectrum discrepancies in deep network generated images,'' 2020. [Online]. Available: \url{https://arxiv.org/abs/1911.06465}
\BIBentrySTDinterwordspacing

\bibitem{corvi2022detectionsyntheticimagesgenerated}
\BIBentryALTinterwordspacing
R.~Corvi, D.~Cozzolino, G.~Zingarini, G.~Poggi, K.~Nagano, and L.~Verdoliva, ``On the detection of synthetic images generated by diffusion models,'' 2022. [Online]. Available: \url{https://arxiv.org/abs/2211.00680}
\BIBentrySTDinterwordspacing

\bibitem{ricker2024detectiondiffusionmodeldeepfakes}
\BIBentryALTinterwordspacing
J.~Ricker, S.~Damm, T.~Holz, and A.~Fischer, ``Towards the detection of diffusion model deepfakes,'' 2024. [Online]. Available: \url{https://arxiv.org/abs/2210.14571}
\BIBentrySTDinterwordspacing

\bibitem{chai2020makesfakeimagesdetectable}
\BIBentryALTinterwordspacing
L.~Chai, D.~Bau, S.-N. Lim, and P.~Isola, ``What makes fake images detectable? understanding properties that generalize,'' 2020. [Online]. Available: \url{https://arxiv.org/abs/2008.10588}
\BIBentrySTDinterwordspacing

\bibitem{Chen2020SSDGANMT}
\BIBentryALTinterwordspacing
Y.~Chen, G.~Li, C.~Jin, S.~Liu, and T.~H. Li, ``Ssd-gan: Measuring the realness in the spatial and spectral domains,'' in \emph{AAAI Conference on Artificial Intelligence}, 2020. [Online]. Available: \url{https://api.semanticscholar.org/CorpusID:228083505}
\BIBentrySTDinterwordspacing

\bibitem{wang2023diffusiondblargescalepromptgallery}
\BIBentryALTinterwordspacing
Z.~J. Wang, E.~Montoya, D.~Munechika, H.~Yang, B.~Hoover, and D.~H. Chau, ``Diffusiondb: A large-scale prompt gallery dataset for text-to-image generative models,'' 2023. [Online]. Available: \url{https://arxiv.org/abs/2210.14896}
\BIBentrySTDinterwordspacing

\bibitem{ojha2024universalfakeimagedetectors}
\BIBentryALTinterwordspacing
U.~Ojha, Y.~Li, and Y.~J. Lee, ``Towards universal fake image detectors that generalize across generative models,'' 2024. [Online]. Available: \url{https://arxiv.org/abs/2302.10174}
\BIBentrySTDinterwordspacing

\bibitem{gragnaniello2021gangeneratedimageseasy}
\BIBentryALTinterwordspacing
D.~Gragnaniello, D.~Cozzolino, F.~Marra, G.~Poggi, and L.~Verdoliva, ``Are gan generated images easy to detect? a critical analysis of the state-of-the-art,'' 2021. [Online]. Available: \url{https://arxiv.org/abs/2104.02617}
\BIBentrySTDinterwordspacing

\bibitem{cozzolino2021universalganimagedetection}
\BIBentryALTinterwordspacing
D.~Cozzolino, D.~Gragnaniello, G.~Poggi, and L.~Verdoliva, ``Towards universal gan image detection,'' 2021. [Online]. Available: \url{https://arxiv.org/abs/2112.12606}
\BIBentrySTDinterwordspacing

\bibitem{tan2024frequencyawaredeepfakedetectionimproving}
\BIBentryALTinterwordspacing
C.~Tan, Y.~Zhao, S.~Wei, G.~Gu, P.~Liu, and Y.~Wei, ``Frequency-aware deepfake detection: Improving generalizability through frequency space learning,'' 2024. [Online]. Available: \url{https://arxiv.org/abs/2403.07240}
\BIBentrySTDinterwordspacing

\bibitem{Zhang2019DetectingAS}
\BIBentryALTinterwordspacing
X.~Zhang, S.~Karaman, and S.-F. Chang, ``Detecting and simulating artifacts in gan fake images,'' \emph{2019 IEEE International Workshop on Information Forensics and Security (WIFS)}, pp. 1--6, 2019. [Online]. Available: \url{https://api.semanticscholar.org/CorpusID:196622700}
\BIBentrySTDinterwordspacing

\bibitem{frank2020leveragingfrequencyanalysisdeep}
\BIBentryALTinterwordspacing
J.~Frank, T.~Eisenhofer, L.~Schönherr, A.~Fischer, D.~Kolossa, and T.~Holz, ``Leveraging frequency analysis for deep fake image recognition,'' 2020. [Online]. Available: \url{https://arxiv.org/abs/2003.08685}
\BIBentrySTDinterwordspacing

\bibitem{durall2020watchupconvolutioncnnbased}
\BIBentryALTinterwordspacing
R.~Durall, M.~Keuper, and J.~Keuper, ``Watch your up-convolution: Cnn based generative deep neural networks are failing to reproduce spectral distributions,'' 2020. [Online]. Available: \url{https://arxiv.org/abs/2003.01826}
\BIBentrySTDinterwordspacing

\bibitem{li2021frequencyawarediscriminativefeaturelearning}
\BIBentryALTinterwordspacing
J.~Li, H.~Xie, J.~Li, Z.~Wang, and Y.~Zhang, ``Frequency-aware discriminative feature learning supervised by single-center loss for face forgery detection,'' 2021. [Online]. Available: \url{https://arxiv.org/abs/2103.09096}
\BIBentrySTDinterwordspacing

\bibitem{jeong2021bihpfbilateralhighpassfilters}
\BIBentryALTinterwordspacing
Y.~Jeong, D.~Kim, S.~Min, S.~Joe, Y.~Gwon, and J.~Choi, ``Bihpf: Bilateral high-pass filters for robust deepfake detection,'' 2021. [Online]. Available: \url{https://arxiv.org/abs/2109.00911}
\BIBentrySTDinterwordspacing

\bibitem{corvi2023intriguingpropertiessyntheticimages}
\BIBentryALTinterwordspacing
R.~Corvi, D.~Cozzolino, G.~Poggi, K.~Nagano, and L.~Verdoliva, ``Intriguing properties of synthetic images: from generative adversarial networks to diffusion models,'' 2023. [Online]. Available: \url{https://arxiv.org/abs/2304.06408}
\BIBentrySTDinterwordspacing

\bibitem{jamini_style}
AstaGuru, ``Jamini roy style,'' \url{https://www.astaguru.com/blogs/jamini-roy---exploring-the-quintessence-of-india%E2%80%99s-folk-art-44}, n.d.

\bibitem{jamini_dataset}
\BIBentryALTinterwordspacing
M23CSA011, ``Jamini roy dataset,'' GitHub repository, 2025. [Online]. Available: \url{https://github.com/M23CSA011/Jamini-Roy-Dataset}
\BIBentrySTDinterwordspacing

\bibitem{artnet}
ArtNet, ``Jamini roy,'' \url{https://www.artnet.com/artists/jamini-roy/}, n.d.

\bibitem{esser2024scalingrectifiedflowtransformers}
\BIBentryALTinterwordspacing
P.~Esser, S.~Kulal, A.~Blattmann, R.~Entezari, J.~Müller, H.~Saini, Y.~Levi, D.~Lorenz, A.~Sauer, F.~Boesel, D.~Podell, T.~Dockhorn, Z.~English, K.~Lacey, A.~Goodwin, Y.~Marek, and R.~Rombach, ``Scaling rectified flow transformers for high-resolution image synthesis,'' 2024. [Online]. Available: \url{https://arxiv.org/abs/2403.03206}
\BIBentrySTDinterwordspacing

\bibitem{diffusion}
\BIBentryALTinterwordspacing
J.~Ho, A.~Jain, and P.~Abbeel, ``Denoising diffusion probabilistic models,'' 2020. [Online]. Available: \url{https://arxiv.org/abs/2006.11239}
\BIBentrySTDinterwordspacing

\bibitem{stability2024stablediffusion3}
S.~{AI}, ``Stable diffusion 3,'' \url{https://huggingface.co/stabilityai/stable-diffusion-3-medium-diffusers}, 2024.

\bibitem{ZhangDenoiser2017}
K.~Zhang, W.~Zuo, Y.~Chen, D.~Meng, and L.~Zhang, ``Beyond a gaussian denoiser: Residual learning of deep cnn for image denoising,'' \emph{IEEE Transactions on Image Processing}, vol.~26, no.~7, pp. 3142--3155, 2017.

\end{thebibliography}

\clearpage

\onecolumn 

\section*{Appendix}


\subsection{Artifact Analysis}

A crucial step in developing more effective detectors is understanding which image features can help distinguish fake images from real ones. In this study, methods from \cite{corvi2023intriguingpropertiessyntheticimages} are adopted to analyze artifacts in the dataset, which includes both real and synthetic Jamini Roy artworks. The synthetic images were generated using a fine-tuned Stable Diffusion model. The analysis is carried out by observing second-order statistics in both the spatial domain, using the image autocorrelation function, and the frequency domain, through the image power spectrum.

To examine structural artifacts, the autocorrelation function is calculated as:

\begin{equation}
R_{x_i}(\Delta m, \Delta n) = \langle x_i(m, n) x_i(m + \Delta m, n + \Delta n) \rangle
\end{equation}
where $x_i(m, n)$ represents the pixel intensity of the $i$-th image at spatial location $(m, n)$, $\Delta m$ and $\Delta n$ denote the spatial displacements between pixel pairs, and $\langle \cdot \rangle$ indicates the spatial average.

\begin{equation}
R_x(\Delta m, \Delta n) = \frac{1}{I} \sum_{i=1}^I R_{x_i}(\Delta m, \Delta n)
\end{equation}
where $I$ is the total number of images in the dataset. This function summarizes the statistical correlation between pixels separated by a $(\Delta m, \Delta n)$ displacement.

The Fourier transform is applied as follows:

\begin{equation}
\mathcal{F}[x_i(m, n)] = \sum_{m=1}^M \sum_{n=1}^N x_i(m, n) e^{-j2\pi\left(\frac{km}{M} + \frac{ln}{N}\right)}
\end{equation} 
where $\mathcal{F}[\cdot]$ denotes the Fourier transform operator and $M$ and $N$ are the height and width of the image in pixels, 

\begin{equation}
X_i(k, l) = \mathcal{F}[x_i(m, n)]
\end{equation} 
where $k$ and $l$ are frequency indices corresponding to horizontal and vertical spatial frequencies.

The power spectrum is given by:

\begin{equation}
S_x(k, l) = \frac{1}{I} \sum_{i=1}^I |X_i(k, l)|^2
\end{equation}
The power spectrum accounts for the fraction of the total image power concentrated at a given $(k/M, l/N)$ frequency pair.

The noise residual is defined as:

\begin{equation}
r_i(m, n) = x_i(m, n) - \mathcal{D}(x_i(m, n); \sigma)
\end{equation}
where $\mathcal{D}(x_i(m, n); \sigma)$ is the denoising function applied to the pixel $x_i(m, n)$, with noise parameter $\sigma = 1$, as per Zhang et al. \cite{ZhangDenoiser2017}. The noise residual $r_i(m, n)$ captures the difference between the original pixel intensity and its denoised counterpart.

By analyzing the central $65 \times 65$ region of the autocorrelation function, distinct artifacts in the synthetic images are revealed, indicating model-specific patterns.

\textbf{Power Spectrum:} Both models display a cross-shaped pattern, indicating regular frequency components. ControlNet and IPAdapter demonstrates better frequency preservation across noise levels, with a more uniform central energy distribution and reduced high-frequency artifacts. In contrast, the base model shows greater spectral variation, particularly at higher noise levels.

\textbf{Autocorrelation:} Both versions show periodic checkerboard artifacts, a sign of synthetic generation. However, ControlNet and IP-Adapter maintains more consistent autocorrelation patterns across noise levels, with better preservation of local structure and reduced artifact prominence at higher noise.

\textbf{Noise Impact:} As noise increases (0.0 to 0.75), both models degrade progressively. ControlNet and IP-Adapter retains structural details better at high noise levels.

\begin{figure*}[h!]
    \centering
    \includegraphics[width=\textwidth]{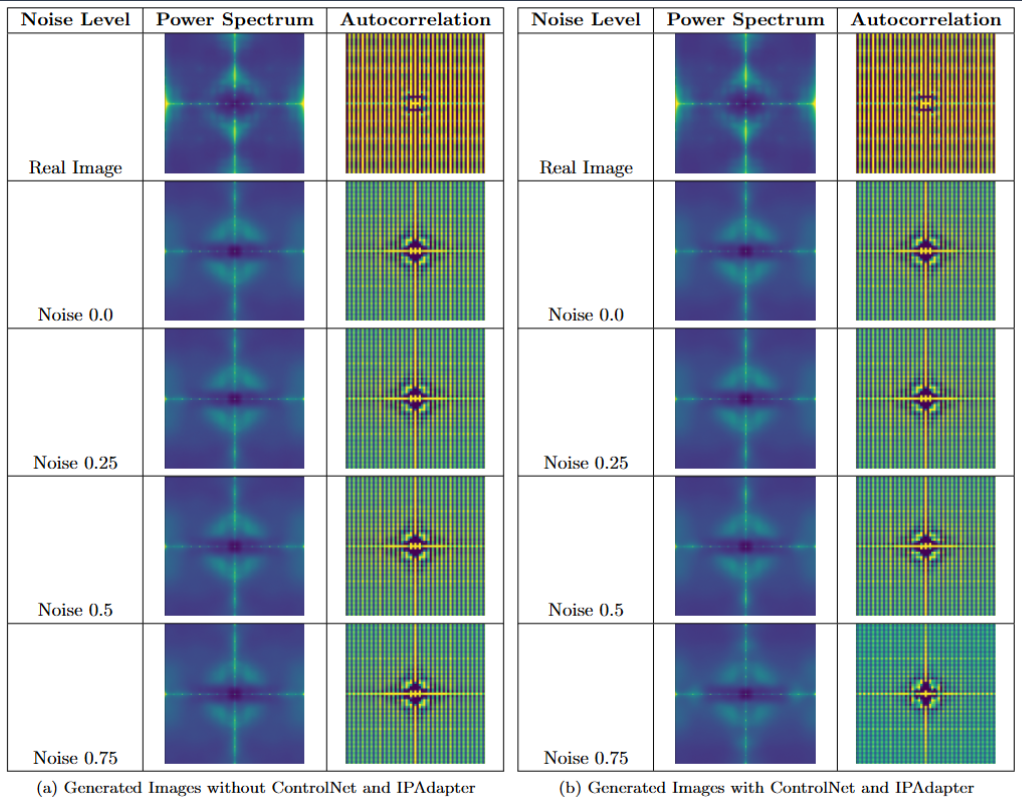}
    \caption{Power Spectrum and Autocorrelation of noise residuals for real and synthetic images generated with and without ControlNet and IPAdapter}
    \label{fig:artifacts_analysis}
\end{figure*}

Hence, the findings showed in figure~\ref{fig:artifacts_analysis} indicate that while synthetic Jamini Roy artworks approximate stylistic elements well, the distribution of autocorrelation and frequency spectra differs in ways that suggest model-specific artifacts, stressing the need for further refinement in deep fake detection for culturally specific datasets like Indian art.

\end{document}